\documentclass[sigconf]{acmart}
\usepackage{amsthm}
\usepackage{algorithmic}
\usepackage[lined,linesnumbered]{algorithm2e}
\usepackage{graphicx}
\usepackage{siunitx}
\usepackage{booktabs}
\usepackage{textcomp}
\usepackage{xcolor}
\usepackage{balance}
\usepackage{subcaption}
\usepackage{url}

\theoremstyle{plain}
\newtheorem{theorem}{Theorem}
\theoremstyle{definition}
\newtheorem{definition}[theorem]{Definition}

\DeclareMathOperator*{\argmax}{argmax}
\DeclareMathOperator*{\argmin}{argmin}

\def\BibTeX{{\rm B\kern-.05em{\sc i\kern-.025em b}\kern-.08em
    T\kern-.1667em\lower.7ex\hbox{E}\kern-.125emX}}

\copyrightyear{2022} 
\acmYear{2022} 
\setcopyright{rightsretained} 
\acmConference[CAIN'22]{1st Conference on AI Engineering - Software Engineering for AI}{May 16--24, 2022}{Pittsburgh, PA, USA}
\acmBooktitle{1st Conference on AI Engineering - Software Engineering for AI (CAIN'22), May 16--24, 2022, Pittsburgh, PA, USA}
\acmDOI{10.1145/3522664.3528606}
\acmISBN{978-1-4503-9275-4/22/05}

\begin{document}

\title{Influence-Driven Data Poisoning in Graph-Based Semi-Supervised Classifiers}

\author{Adriano Franci}
\affiliation{%
  \institution{SnT, Université de Luxembourg}
      \country{Luxembourg}
}
\email{adriano.franci@uni.lu}
\author{Maxime Cordy}
\affiliation{%
  \institution{SnT, Université de Luxembourg}
      \country{Luxembourg}
}
\email{maxime.cordy@uni.lu}
\author{Martin Gubri}
\affiliation{%
  \institution{SnT, Université de Luxembourg}
      \country{Luxembourg}
}
\email{martin.gubri@uni.lu}
\author{Mike Papadakis}
\affiliation{%
  \institution{SnT, Université de Luxembourg}
    \country{Luxembourg}
}
\email{michail.papadakis@uni.lu}
\author{Yves Le Traon}
\affiliation{%
  \institution{SnT, Université de Luxembourg}
  \country{Luxembourg}
}
\email{yves.letraon@uni.lu}

\begin{abstract}
Graph-based Semi-Supervised Learning (GSSL) is a practical solution to learn from a limited amount of labelled data together with a vast amount of unlabelled data. However, due to their reliance on the known labels to infer the unknown labels, these algorithms are sensitive to data quality. It is therefore essential to study the potential threats related to the labelled data, more specifically, label poisoning. In this paper, we propose a novel data poisoning method which efficiently approximates the result of label inference to identify the inputs which, if poisoned, would produce the highest number of incorrectly inferred labels. We extensively evaluate our approach on three classification problems under 24 different experimental settings each. Compared to the state of the art, our influence-driven attack produces an average increase of error rate 50\% higher, while being faster by multiple orders of magnitude.  Moreover, our method can inform engineers of inputs that deserve investigation (relabelling them) before training the learning model. We show that relabelling one-third of the poisoned inputs (selected based on their influence) reduces the poisoning effect by 50\%.
\end{abstract}

\maketitle  
\section{Introduction}

Recent advances in Machine Learning (ML) resulted in an unprecedented interest towards embedding such technologies in software systems. To be effective though, ML algorithms generally require a massive amount of data together with their expected prediction outcomes (labels). Such labelling activities are expensive and time-consuming as they are typically performed manually by humans. Thus, acquiring labelled data is seen as a major obstacle to the widespread adoption of ML.

\begin{figure*}
    \centering
    \includegraphics[width=0.85\linewidth]{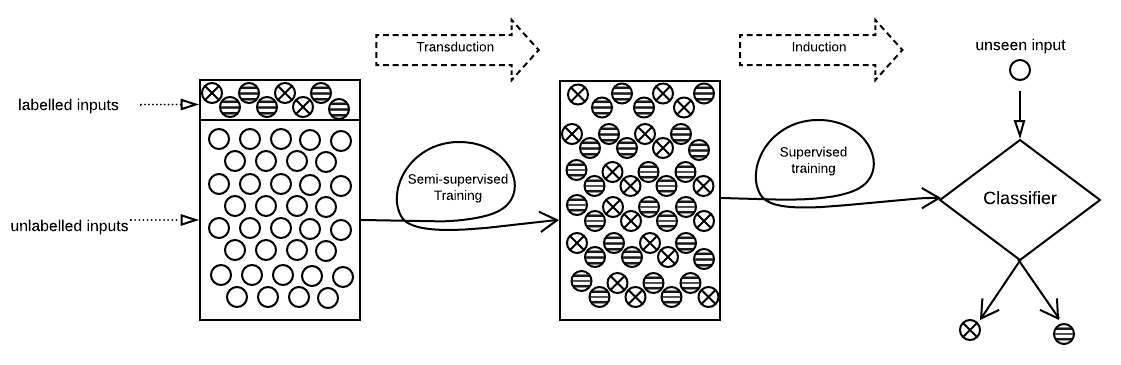}
    \caption{Principles of semi-supervised learning. During transductive learning, the labels of the unlabelled data are inferred from the known labels of the labelled inputs. During inductive learning, the whole data set (with known and inferred labels) are used to train a supervised model, which can be used to classify previously unseen data.}
    \label{fig:ssl}
\end{figure*}

To alleviate this problem, Semi-Supervised Learning (SSL) algorithms \cite{Zhu2005} were proposed to exploit a limited amount of labelled data together with a vast amount of unlabelled data (whose acquisition is generally inexpensive). Their principle (see Figure~\ref{fig:ssl}) is to use the labelled inputs to infer the correct label of unlabelled inputs that are similar (i.e., close in the feature space). Such solutions are flexible: they can be used  for both \emph{transductive} learning (i.e., inferring the labels of the unlabelled inputs) and \emph{inductive} learning (i.e., using the initial labelled data and the inferred labels to train a supervised model and make predictions on unseen data). Moreover, SSL was shown to produce significant improvements over ``fully'' supervised algorithms (which learn from labelled data only) and unsupervised ones (which do not use data labels) \cite{kaariainen2005,ben2008}. It has been successfully applied in a variety of domains including image classification \cite{Fergus2009,Iscen2019}, drug interaction discovery \cite{Zhang2015} and social media mining \cite{speriosu2011}.

A major thrust in SSL comes from its greater reliance on data quality to perform correct predictions \cite{Li2019}. This makes such algorithms vulnerable to data poisoning \cite{liu2019}, i.e. intentional or unintentional alterations of training data that mislead the learning of prediction models (e.g., reducing prediction accuracy). Data poisoning is considered a serious threat to ML-enabled systems \cite{Joseph2013} that is prominent in cases where the data acquisition process is not entirely trustworthy. This phenomenon has been observed and studied in various contexts such as intrusion detection \cite{Perdisci2006,Cordy2019}, face recognition \cite{Biggio2013} and crowdsourcing systems \cite{Wang2014,TAY2016}. While much research studied data poisoning in supervised models (see, e.g., \cite{Mei2015,Xiao2015}), data poisoning in SSL remains largely unexplored: in their NeurIPS'19 paper, Liu et al. proposed the first and only data poisoning framework for SSL \cite{liu2019}. Studying this security aspect is not only important to identify potential threats on safety- and business-critical software systems relying on SSL, but it is also a prerequisite to design appropriate countermeasures.

Existing studies typically rely on dedicated algorithms -- named \emph{data poisoning attacks} -- that aim to find the data alterations with the maximal impact on SSL. Thanks to these algorithms, researchers and engineers can estimate the impact that the worst-case data alterations will have on the target ML-enabled systems. In turn, they can design countermeasures to best protect the systems against data poisoning instances.

In this paper, we propose a novel data poisoning attack for SSL which is simple to compute (involves common matrix operations only), effective (in reducing accuracy) and efficient (requires low computation costs). Our attack emulates a common data poisoning scenario where some inputs have received an incorrect label before learning occurs (label poisoning) \cite{Paudice2018}. Such incorrect labels, in turn, alter the inference capabilities of learning models. The key idea of our method is to exploit mathematical properties of SSL to identify ``weak spots'' in the labelled data set, i.e. inputs that, if mislabelled, would yield the largest number of wrongly-inferred labels. To do so, we define a metric to approximate the \emph{influence} of labelled inputs onto unlabelled ones. Based on the proposed metric, we can rank labelled inputs wrt. the expected impact of poisoning their label. Inputs ranked higher are the ideal target of attacks since their poisoning yields the highest reduction in prediction accuracy. Hence, we refer to our attack as 
\emph{influence driven}.

We extensively evaluate our approach on three classification problems from different application domains (image recognition and text classification), in both transductive learning settings and inductive learning settings. As a first step, we empirically demonstrate that our influence metric has medium-to-strong correlations (between 0.31 and 0.99) with the error rate resulting from poisoning the labelled inputs. This indicates that the metric captures well the expected impact of poisoning each of the inputs and, therefore, is convenient to prioritize them. Secondly, we show that our metric remains effective (i) when used to select multiple inputs to poison and (ii) during inductive learning, i.e. when the labelled data resulting from SSL are used to train a supervised model to classify unseen data. Compared to the state of the art, our poisoning attack yields a significantly higher number of new misclassifications than the state of the art (50\% higher on average), while being 2 to 3 times faster. Thus, it forms a new, strong baseline for future research in SSL data poisoning.

Finally, we investigate the use case where our influence metric supports engineers in identifying the most critical inputs to relabel (re-investigate the correct label before learning) in order to protect their system against the effect of label poisoning. We show that ensuring the label correctness of the most influential inputs more than halves the error rate on average, whereas labelling additional inputs brings minor benefits (less than 15\% of average error rate reduction). 

To summarize, our contribution is four-fold:
\begin{enumerate}
    \item We formalise the concept of input influence in SSL and exploit it to identify the most influential labelled inputs during transductive learning (label inference for unlabelled data). 
    \item We exploit this concept to design a novel data poisoning attack, which consists of altering the label of the most influential inputs.
    \item We conduct an in-depth experimental study showing the benefits of our approach. Our results indicate that our attack transfers well to different SSL algorithms and to inductive learning settings (using supervised models). Compared to the state of the art \cite{liu2019}, our method is significantly more effective (the average increase in model error rate is 50\% higher on average) and more efficient (2 to 74 times faster).
    \item We show that an effective strategy to counter data poisoning is to relabel the most influential inputs, guaranteeing their correctness. Compared to labelling an equivalent number of additional inputs, this is the most effective strategy to alleviate the effect of our attack. 
\end{enumerate}

\section{Background and Related Work}

Semi-supervised learning \cite{Zhu2005} is a particular form of machine learning that attempts to maximize the benefits of learning from a limited amount of labelled data together with a vast amount of unlabelled data. It contrasts with supervised learning which necessitates all training data to be labelled, and unsupervised learning which does not rely on labels but only exploit statistical relations between the input (e.g., to form clusters). 

Among the different families of SSL algorithms (see \cite{Zhu2005} for an overview), graph-based methods are the most popular because they are effective (in inferring unknown labels from labelled data), efficient (in computation time) and straightforward to implement (based on common matrix operations). The two established graph-based SSL algorithms are label propagation \cite{Zhu2002} and label spreading \cite{Zhou2004}. Both consist of computing the label likelihood of any unlabelled input based on the labels of its neighbouring inputs. This computation follows an iterative process until either the label likelihood values converge or a predefined number of iterations is reached. The key differences between the two algorithms lie in how they compute input similarity and how they propagate the likelihood values from one iteration to the other.

Research on ML security is mainly spearheaded by adversarial machine learning. One distinguishes two types of adversarial attacks~\cite{Pitropakis2019}: \textit{evasion attacks} and \textit{poisoning attacks}.

Evasion attacks occur at prediction (test) time. They consist of perturbating the input sent to the ML model (yielding an ``adversarial input'') to cause misclassification. These attacks have been studied intensively over the recent years \cite{Pitropakis2019}. They are also commonly used to improve the thoroughness of ML system test suites by generating failure-inducing inputs \cite{Zhang2019}.

By contrast, poisoning attacks strike at \emph{training time} and aim to degrade the overall performance (e.g., the accuracy) of the ML model \cite{Biggio2012}. These attacks often consist of either \emph{modifying the training data} (falsifying them) or \emph{injecting new data} (containing intentional mistakes) into the training set. In the former case, the attacker can either modify (selected) features of the poisoned inputs (feature poisoning) or their associated label (label poisoning). Although it has been less studied, the third type of poisoning attack is \emph{algorithm corruption}, during which the attacker alters the logic of the training process, thereby changing the way the ML model learns \cite{gu2019}. The new attack we design in this work belongs to the label poisoning category, which requires fewer assumptions about the attacker capability (e.g., it does not require any knowledge about the features of the inputs or access to the internal structure of the training algorithm).

Data poisoning in SSL has been scarcely studied. To the best of our knowledge, the only approach has been recently designed by Liu et al. \cite{liu2019}. While it supports both regression and binary classification problems, we focus here on classification, which they deem to be the most challenging setting \cite{liu2019}. Like ours, their method leans on the properties of label propagation. It consists of a search algorithm that selects the labelled inputs to poison to maximize the expected increase in error rate, computed as the difference between the ground truth labels and an approximation of the labels that will be inferred. Thus, their method assumes that the attacker has access to the real labels of the unlabelled inputs or can compute them reliably using a surrogate SSL model. We do not make such assumption and select the labelled inputs to be poisoned based only on their relative influence on the unlabelled inputs (without knowing the truth labels). Our evaluation compares their methods with ours on various experimental settings covering different factors like proportion of labelled inputs, poison budget, dataset and used learning algorithms.

\section{Problem Formulation}

\subsection{SSL and Performance Metrics}

We consider a classification problem where $C = \{c_1, \dots, c_k\}$ is the set of classes. Let $X = (X_L \cup X_U) \subseteq \mathbb{R}^D$ be a set of inputs represented by $D$-dimensional feature vectors, such that $X$ is divided into a set $X_L$ of $l$ labelled inputs and a set $X_U$ of $u$ unlabelled inputs. We arbitrarily index those inputs such that the labelled ones are placed first. That is, $X_L = \{x_1, \dots, x_l\}$ and $X_U = \{x_{l+1}, \dots , x_{l+u}\}$. Finally, we denote by $\mathbf{y}_L=(y_1, \dots, y_l) \in C^l$ the label vector associated to $x_1, \dots, x_l$, respectively. Then, the goal of SSL is to infer the label vector $\mathbf{y}_U = (y_{l+1},\dots, y_{l+u})$ of $X_U$ given $X$ and $\mathbf{y}_L$, that is, to learn a function $f(x \in X_U | X, \mathbf{y}_L) \in C$ whose output should be as close as possible to the correct label vector $\mathbf{y}^*_U = (y^*_{l+1}, \dots, y^*_{l+u})$.

One can measure the performance of SSL according to two different goals. If the goal is transductive learning (i.e., to infer the labels $\mathbf{y}_U$ correctly), then performance can be measured by the \emph{transductive accuracy} of the SSL algorithm, defined as 
$$\frac{\sum_{x_i \in X_U} \mathbf{1}_{\{f(x_i) = y^*_i\}}}{|X_U|}$$
where $\mathbf{1}_{\{a = b\}}$ = 1 if $a = b$ and 0 otherwise.

The second goal is \emph{inductive learning}. It aims to build an arbitrary supervised model $\mathcal{M}$ that generalizes to some unseen data $X_G$. In such a case, $\mathcal{M}$ uses as a training set both the labelled inputs with their known labels and the unlabelled inputs with their labels inferred by SSL. Performance can then be measured by the \emph{inductive accuracy}, i.e.,
the percentage of unseen data that $\mathcal{M}$ classifies correctly. In practice, inductive accuracy is approximated by measuring the accuracy of $\mathcal{M}$ on a carefully collected representative subset $X_{test} \subset X_G$ of the unseen data, named the \emph{test set}.

\subsection{Threat Model}

We define an attack as a process that aims to reduce the performance of SSL (transductive or inductive accuracy) by poisoning available data. More precisely, we consider label poisoning attacks where the attack can alter the known labels $\mathbf{y}_L$.

\paragraph{Attack goal} The attack aims to maximize the error rate of the classifier. We target both transductive learning (where the end goal of the system is to correctly infer the label of unlabelled data) 
and inductive learning (where known and inferred labels are used to train a supervised model). 
Thus, the achievements of the attack can be measured by the error rate (i.e., 100\% minus the accuracy) of the SSL algorithm or the supervised model, respectively, after poisoning. For simplicity, we restrict the scope to binary classification. The attack is indiscriminate as it does not target specific inputs and aims for generic errors (it does not target particular classes). 

\paragraph{Attack knowledge} The attack has access to the known dataset ($X$ and $\mathbf{y}^*_L$). However, our attack is black-box as it has knowledge about neither the graph-based SSL algorithm used during transduction nor the supervised models used for induction.

\paragraph{Attack capability} The attack can alter any label of the labelled inputs and can do so for a limited number $k$ of labels (to minimize its actions). The features of the inputs cannot be altered. 

\paragraph{Objective function} In summary, the objective of the attack is to find a perturbation vector $\delta^* \in (C \cup \{0\})^l$ under the constraint $|\{j | \delta^*_j \neq 0\}| = k$,  such that altering $\mathbf{y}_L$ into $\mathbf{y}_L \oplus \delta^*$ (where $(\mathbf{y}_L \oplus \delta^*)_{j} = (\mathbf{y}_L)_j$ if $\delta^*_j = 0$; and $(\mathbf{y}_L \oplus \delta^*)_{j} = \delta^*_j$ otherwise) results in the maximal (transductive or inductive) error rate.

\section{Influence-Driven Data Poisoning}

\subsection{Preliminaries}
\label{sec:approach-preliminaries}

Graph-based SSL is a popular family of SSL algorithms that were shown to reach high accuracy at affordable computation costs~\cite{Seeger2000}. They consist of building a fully connected graph where vertices are (labelled and unlabelled) inputs and where edges are weighted in proportion to the similarity between inputs. Then, SSL uses this representation to infer the label of any unlabelled input based on the weight of its edges.

\begin{figure}
    \centering
    \subfloat[Initial situation]{\includegraphics[width=0.49\linewidth]{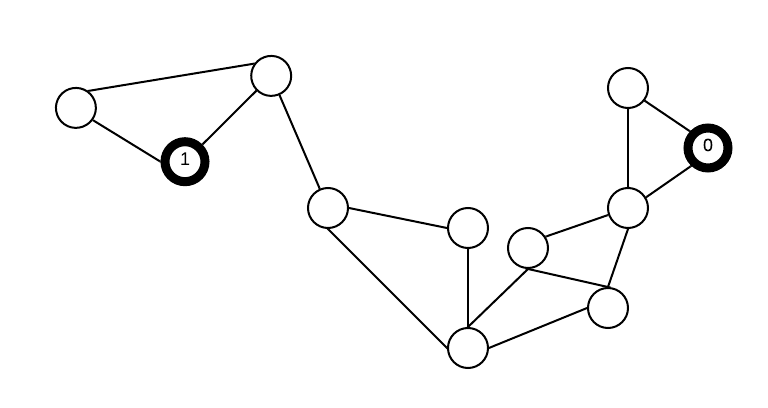}}
    \hfill
    \subfloat[Energy propagation]{\includegraphics[width=0.49\linewidth]{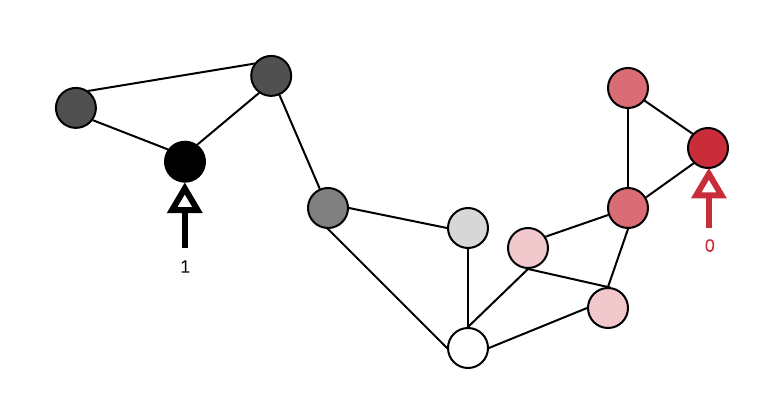}}
    \caption{Energy propagation principle used by label propagation to infer the labels of the unlabelled inputs (light circles) from labelled inputs (tick circles with class number inside). Each colour represents a distinct class, while color gradient represents the influence strength of the propagated class label. This influence diminishes as there is more distance between the original labelled input and the unlabelled input.}
    \label{fig:label propagation}
\end{figure}

Label propagation \cite{Zhu2002,Zhu2003} is one of the most popular graph-based SSL algorithms. As illustrated in Figure \ref{fig:label propagation}, 
it utilizes the \textit{energy propagation principle} to achieve transductive learning. Intuitively, any labelled input emits its label towards any unlabelled input to influence its labelling, such that this influence (the 
``label energy'') fades away as the distance between the two inputs increases. Then, the label of any unlabelled input is inferred from the sum of the influences it receives (both from known labels and other inferred labels). Inferred labels also propagate their own energy. Thus, label propagation is an iterative process that stops after inferred labels have reached a steady state.

Hence, label propagation starts by computing a similarity distance between any pair of inputs. A popular class of functions to measure similarity between feature vectors are the Kernel functions. In particular, label propagation often uses the  \emph{Radial Based Function kernel} (RBF kernel)~\cite{Scholkopf2000}. Accordingly, the similarity $w_{ij}$ between $x_i$ and $x_j$ is given by $w_{ij} = \texttt{exp}(-\gamma||x_i-x_j||^2) $ with $\gamma = (2 \sigma 
^2)^{-1}$. For convenience, we define the $(l+u) \times (l+u)$ weight matrix $W =\begin{bmatrix}
{W}_{LL} & {W}_{LU} \\
    {W}_{UL} & {W}_{UU}
\end{bmatrix}$ 
such that $W_{ij} = w_{ij}$. The unknown labels are then predicted via the energy minimization principle:
$$\mathbf{y} = \argmin_{\hat{y}} \sum_{i,j} W_{ij} (\hat{y_i} - \hat{y_j})^2 = \hat{y}^T (\mathbf{D} - W) \hat{y}, \text{ s.t. } \hat{y_{:l}} = \mathbf{y}_L $$ where $\mathbf{D} = diag_{\sum_k W_{ik}}$. This problem admits a closed form solution: 
$$\mathbf{y}_U = (D_{[(l+1):(l+u),(l+1):(l+u)]} - W_{UU})^{-1} W_{UL} \mathbf{y}_L.$$
Let $T = \begin{bmatrix}
{T}_{LL} & {T}_{LU} \\
    {T}_{UL} & {T}_{UU}
\end{bmatrix}$ be the  
$(l+u)\times (l+u)$ label transition matrix defined as
$$T_{ij}= \left \lbrace 
    \begin{array}{ll}
     1, & i \leq l \land i = j;\\
     0, & i \leq l \land i \neq j ;\\
     \frac{w_{ij}}{\sum_{k=1}^{l+u} w_{kj}}, & l < i < l + u.\\
    \end{array}
\right . $$ 
and $\bar{T}$ the matrix obtained by row-normalizing $T$. Then the above solution can be rewritten as:
$$\mathbf{y}_U = (I - \bar{T}_{UU})^{-1} \bar{T}_{UL} \mathbf{y}_L$$
which can be computed as the fixed point of an iterative algorithm:
\begin{enumerate}
    \item $Y \leftarrow \bar{T} \cdot Y$ 
    \item $Y \leftarrow \begin{bmatrix} Y^{0}_{L} & {Y}_{U} \end{bmatrix}^{\top}$
\end{enumerate}
where $Y^{0}_L$ denotes the initial submatrix of $Y$ corresponding to the known (clamped) labels.

\begin{figure}
\subfloat[Before inference, w/o poison]{\includegraphics[width=0.48\linewidth]{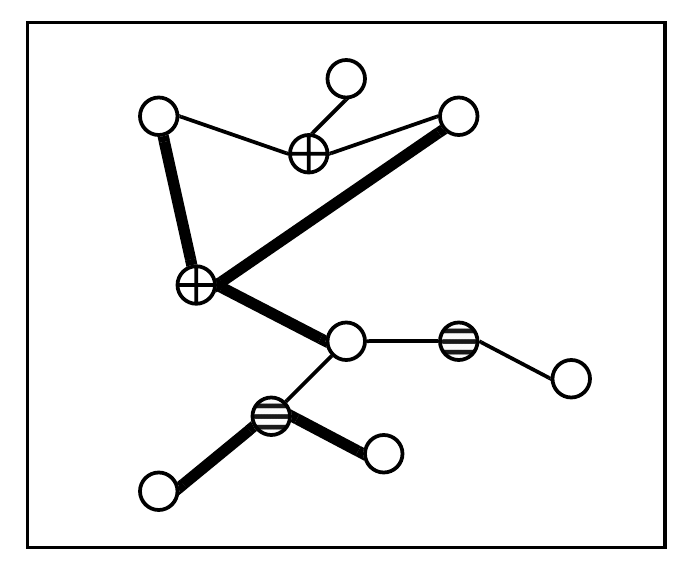}}
\hfill
\subfloat[Label inference result]{\includegraphics[width=0.48\linewidth]{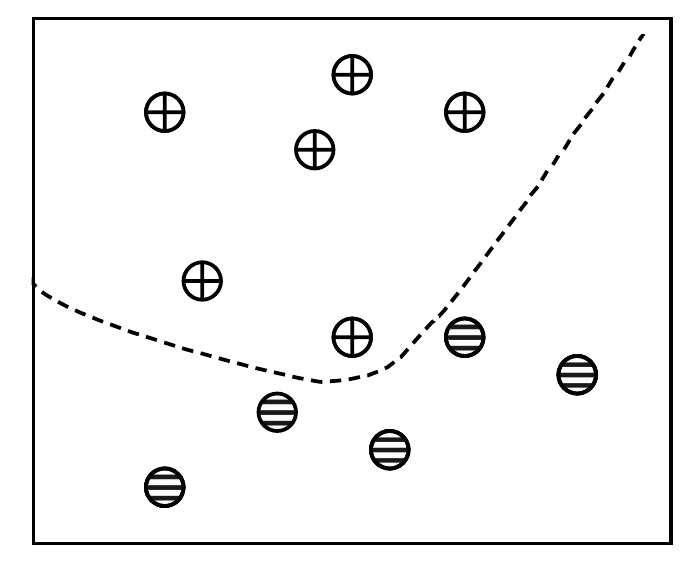}}
\\
\subfloat[Before inference, with poison]{\includegraphics[width=0.48\linewidth]{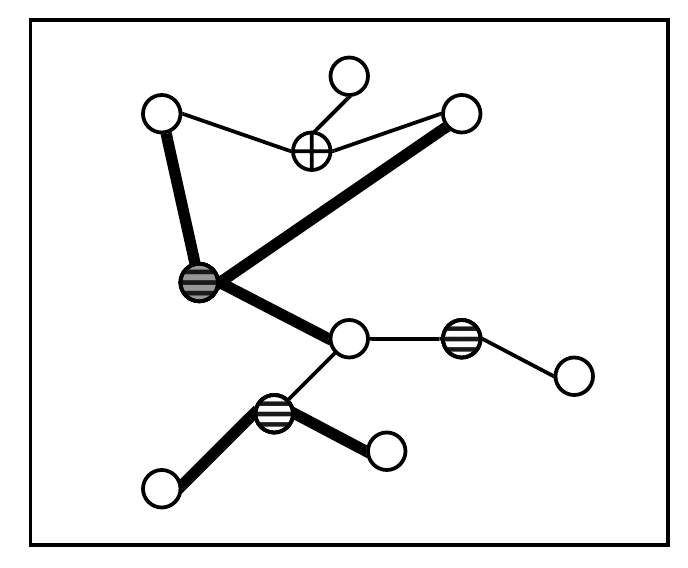}}
\hfill
\subfloat[Label inference result]{\includegraphics[width=0.48\linewidth]{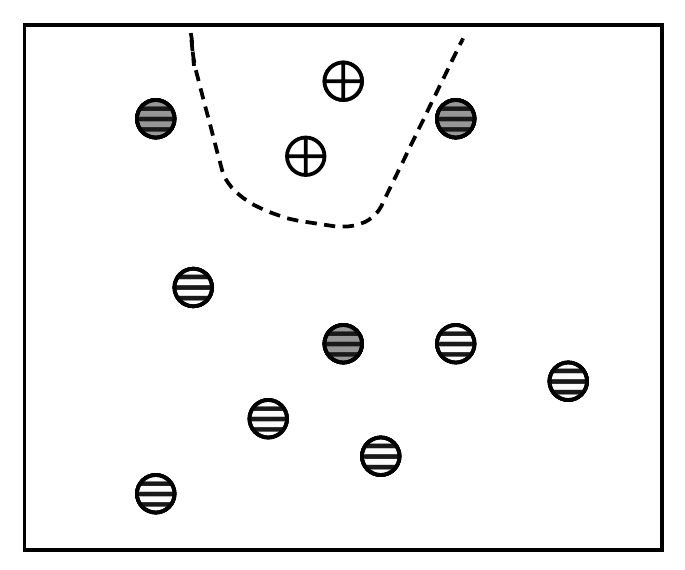}}

\caption{Principles of our influence-based poisoning attack.}
\label{fig:poisonIntuition}
\end{figure}

\subsection{Influence-driven Poisoning Attack}
We show in Figure \ref{fig:poisonIntuition} the principles of our influence-based poisoning attack. The left figure shows the influence graph between inputs (both labelled -- filled -- and unlabelled -- empty -- inputs), where a ticker edge indicates a major influence (no edge means infinitesimal influence). Right figures show the result of label inference. The key idea of our approach is to identify the labelled inputs that have a major influence over a maximum number of unlabelled inputs, such that poisoning their label maximizes the number of wrong inferences. Here, the greyed input was the most influential and selected as the target to poison. This resulted in three labels inferred incorrectly.

To build our method, we associate matrix $\bar{T}$ with a probabilistic interpretation: $\bar{T}{ij}$ is the estimated probability that $x_i$ takes its label from $x_j$. Remember that $\bar{T}_{ij}$ takes into account both the spread of all inputs' label energy and their competitive influence onto determining the inferred labels. In other words, $\bar{T}_{ij}$ measures the relative \emph{influence} of $x_j$ onto $x_i$ (compared to the other inputs). By restricting $j$ to labelled inputs, we obtain the direct influence of any labelled input $x_j$ onto any unlabelled input $x_i$.

\begin{definition}
Let $x_i \in X_U$ and $x_j \in X_L$. The \emph{direct influence} of $j$ onto $i$, noted $e_{ij}$, is given by 
$e_{ij} = \bar{T}_{ij}$.
\end{definition}
The above influence metric disregards any indirect influence of $x_j$ onto $x_i$, i.e. the energy received by $x_i$ which transits from $x_j$ to any intermediary input $x_k$ before reaching $x_i$. This means that $e_{ij}$ is only an approximation of the total influence of $x_j$ onto $x_i$, which would be obtained by repeatedly multiplying $\bar{T}$ by itself until it converges, that is by computing $\lim_{n \rightarrow \infty} \bar{T}^{n}$. Thus, the advantage of our direct influence metric is to avoid those costly computations while remaining a good approximation of the total influence (as later revealed by our evaluation results in Section \ref{sec:results}).

Leaning on the direct influence metric, we are able to identify which labelled input has the highest direct influence onto a given unlabelled input.
\begin{definition}
\label{def:most-influential}
The \emph{most influential} input of $x_i \in X_U$ is given by 
$$s^*_i = \argmax_{x_j \in X_L} e_{ij}$$
\end{definition}

Thus, $s^*_i$ is the labelled input such that altering its label has the highest likelihood (amongst all labelled inputs) to change the inferred label of $x_i$. In particular, if $s^*_i$ holds more than half of the influence towards $i$, then poisoning its labels guarantees that the label of $i$ is inferred incorrectly.\footnote{Actually, this may not be the case if indirect influences reduce the relative influence of $s^*_i$ onto $x_i$. The key idea of our approach is to heuristically ignore the indirect influence between inputs to reduce the computation time of the influence metrics. Hence, it inherently remains an approximation.} 

\begin{definition}
\label{def:most-influential}
Let $x_i \in X_U$ and $s^*_i$ be the most influential input of $x_i$. Then $s^*_i$ is a major influencer of $x_i$ iff
$$\frac{e_{is^*_i}}{\sum_{j\in X_L}e_{ij}} > 0.5$$
\end{definition}

By generalizing this concept to the whole set of unlabelled inputs $X_U$, we can compute the number of unlabelled inputs for which a given labelled input is a major influencer.

\begin{definition}
Let $X_U$ be the set of unlabelled inputs and $x_j \in X_L$ be a labelled input. The \emph{Major Influence Range} (MIR) of $x_j$ w.r.t. $X_U$ is given by
$$MIR(x_j) = |\{x_i \in X_U \text{ s.t. } \frac{e_{ij}}{\sum_{x_k\in X_L}e_{ik}} > 0.5\}|$$
\end{definition}
Thus, $MIR(x_j)$ captures the number of unlabelled inputs that would receive an incorrect label if the label of $x_j$ was poisoned. This paves the way to design our \emph{influence-driven data poisoning attack}: given a budget of $k$ labels to poison, our attack alters the label of the $k$ labelled inputs with the highest MIR.

Overall, our approach only requires computing the matrix $\bar{T}$ and extracting the MIR value of each labelled input from it. Thus, its worst-case time complexity is $\mathcal{O}((l+u)^2)$.

\section{Experimental Protocol}

\subsection{Methodology and Research Questions}

Our data poisoning method relies on the assumption that the MIR of any input captures well the number of incorrectly inferred labels that would result from poisoning that input. Accordingly, our first step is to validate this hypothesis. This is important to ensure that poisoning inputs ranked higher (according to their MIR) yields a higher number of incorrectly inferred labels. We ask:

\begin{description}
    \item[\textbf{RQ1}] \hspace{0.05cm} \emph{Can the MIR metric identify the inputs whose poisoning yields the highest error rate?}
\end{description}
To answer this question, we measure the correlation, over the whole labelled input set, between the MIR of any input and the error rate of transductive learning after altering the label of that input. Thus, given $l$ labelled inputs, we apply label propagation $l$ times (once per poisoned labelled input). 

To measure the correlation, we use the Kendall coefficient because, as an ordinal association metric, it focuses on how well MIR ranks the most impactful inputs first (irrespective of the actual error rates). Thus, if one has to select a limited number of inputs to poison, one should select the inputs of higher ranking. To complement our analysis we also use the Pearson's correlation coefficient, which captures linear relationships between two variables (here, MIR and the resulting error rate). Thus, a high Pearson correlation would indicate that MIR can also estimate the relative impacts of poisoning those inputs (taking into account the induced error rate).

As the MIR metric calculation lies on the same principle used by the label propagation algorithm, we expect that it performs well on this particular SSL algorithm. To assess the generalization potential of our approach, we repeat the experiments using a second SSL algorithm, label spreading, which uses a different Laplacian matrix to represent the input graph. Again, we report the resulting correlation coefficients. Having the same level of correlation would indicate that our method to select samples to poison transfers well to SSL algorithm that uses a different learning procedure.


Our next question assesses the practical effects that MIR-based data poisoning can have on inductive learning, that is, when supervised classification models are trained with both (known) labelled inputs and inputs whose labelled was inferred by semi-supervised learning. Thus, we ask:
\begin{description}
    \item[\textbf{RQ2}] \hspace{0.05cm} \emph{How effective is MIR-based label poisoning in increasing the error rate of inductive learning?}
\end{description}
We consider the use case where an attacker aims to achieve the maximal effect (increase of error rate) within a limited budget of input labels to poison. Thus, given such a budget $k$, we study by how much the error rate of the supervised model on an independent test set increases after poisoning the top-$k$ inputs ranked by MIR. 
Starting from the set of labelled inputs, we poison $k$ labels before running an SSL algorithm (label propagation or label spreading) to infer the labels of the unlabelled data. Then, using the whole training set (labelled data and unlabelled data with inferred labels), we train a supervised model (defined independently of our poisoning attack) and compute its error rate on unseen test data (which have no intersection with the training data). 

Following this, we compare our method to the state-of-the-art data poisoning attack for SSL, i.e. the two methods proposed by Liu et al. \cite{liu2019}, both in terms of effectiveness (how much it increases error rate) and efficiency (computation time). We ask:
\begin{description}
    \item[\textbf{RQ3}] \hspace{0.05cm} \emph{How does our approach compare to the state of the art data poisoning attack?}
\end{description}
We compare the effectiveness of Liu et al.'s method and ours on both transductive learning and inductive learning. In each case, we record the error rates that result from poisoning the inputs suggested by each method. To compare both methods on a fair ground, we use label propagation to perform transductive learning, since both methods lean on the mathematical properties of this SSL algorithm. 

Having shown that our method constitutes an effective attack, we aim to determine if we can use the same principle of influence to guide engineers in setting up effective countermeasures. Thus, we ask:
\begin{description}
     \item[\textbf{RQ4}] \hspace{0.05cm} \emph{Can MIR drive the design of countermeasures to reduce the effect of label poisoning?}
\end{description}
To answer this question, we simulate a scenario where engineers \emph{relabel} the most influential inputs before transductive learning. Relabelling is indeed a common countermeasure to label poisoning \cite{Paudice2018}. We compare this with the alternative countermeasure of labelling additional inputs to offset the poisoning effect. Such comparison will reveal the most effective allocation of the engineers' effort to reduce the effects of label poisoning.


Here it must be noted that the above are general settings common to all RQs we investigate. Specific and detailed settings required to answer each RQ are given at the beginning of the dedicated sections answering them. 

\subsection{Test Subjects}

\begin{table}
    \centering
    \begin{tabular}{l|r|r|r}
    \textbf{Name} & $|X|$ & $|X_{test}|$ & \# features\\
    \hline
         MNIST (1,7) & 13,007 & 2,163  & 784 \\
         CIFAR-10 (cat, ship) & 10,000 & 2,000 & 3072 \\
         rcv1 & 20,242 & 677,399 & 47,236 \\ 
    \end{tabular}
    \caption{Characteristics of the datasets we use in our experiments. $X$ is the set of (labelled and unlabelled) inputs used during transductive learning. $X_{test}$ is the independent set of unseen data used for testing the inductive learning.}
    \label{tab:datasets}
\end{table}

We run our experiments on three datasets involving two image classification problems (\texttt{MNIST} \cite{lecun1998gradient} and \texttt{CIFAR-10} \cite{CIFAR10}) and one text classification problem (\texttt{rcv1} \cite{Lewis2004}). \texttt{MNIST} and \texttt{CIFAR-10} are widely used in research and considered as a good baseline to observe key trends, in addition to requiring affordable computation cost. Using both brings variety within the same application domain: \texttt{MNIST} is about black-and-white images of handwritten digits while \texttt{CIFAR-10} includes colour images of animals and vehicles. Due to this difference, the different class manifolds in \texttt{MNIST} are closer in the feature space than in \texttt{CIFAR-10}. The last dataset, \texttt{rcv1}, allows observing results in another application domain and on a more challenging dataset (computation wise). It involves inputs with 47,236 features, which drastically increases the time to run our experiments.

It is to be noted that these three datasets involve 10 classes (\texttt{MNIST} and \texttt{CIFAR-10}) and 103 classes (\texttt{rcv1}). 
For simplicity we restrict the scope of our empirical study to binary classification.
We use binary variants of the datasets. For \texttt{MNIST}, we keep the images showing digits 1 and 7, following the protocol of \cite{liu2019}. For \texttt{CIFAR-10}, we keep the classes \texttt{ship} and \texttt{cat}. For \texttt{rcv1}, we directly reuse the binary variant of the dataset independently proposed in the LIBSVM collection of datasets \cite{LIBSVMdata}. The characteristics of the resulting datasets are summarized in Table~\ref{tab:datasets}. 


To perform transductive learning, we consider label propagation and label spreading with the default parameters of their scikit-learn implementation. Both label propagation and label spreading use an RBF kernel, a convergence threshold of $10^{-3}$ and a maximum of 30 iterations. Label spreading also uses a clamping factor of $0.1$.

When our experiments involve inductive learning, we use two supervised models: a random forest and a multilayer perceptron. The random forest is composed of 100 trees, uses Gini impurity as the quality split criterion and considers up to $\sqrt{\# features}$ when looking for the best split. The Multilayer Perceptron is composed of an input layer with \textit{\# features} neurons, a hidden layer with 128 neurons activated by a rectifier function and an output layer with one neuron using a sigmoid activation function.

We chose those two supervised learning methods as their learning process significantly differs from that of graph-based SSL. 
This allows us to observe how well poisoning transfers to models based on different learning principles. 

\subsection{Implementation and hardware}

We implemented our methods in a prototype tool on top of Python 3.7.0. Our tool is open source and publicly available together with our datasets and results.\footnote{\url{https://anonymous.4open.science/r/adversarial_label_propagation-FC83}} As for label propagation and label spreading, we rely on their implementation in the open-source library \textit{scikit-learn}\footnote{\url{https://github.com/scikit-learn/scikit-learn}} \cite{scikit-learn}. The implementation of the supervised models we use (viz. random forest) is also based on \textit{scikit-learn}.

To compare with the state-of-the-art method \cite{liu2019}, we reuse the original implementation provided by the authors.\footnote{\url{https://github.com/xuanqing94/AdvSSL}} It includes two variants of the method: one is deterministic and the other is stochastic. 

All experiments were run on Google Cloud using a virtual machine with 12 VCPU Intel Xeon Skylake 2.0 GHz and 45GB of RAM. Running once all experiments of all RQs required approximately 15 days of computation.
\section{Detailed Setup and Results}
\label{sec:results}

\subsection{RQ1: Correlation with Transductive Learning Error Rate}


\subsubsection{Detailed Setup} 

We measure the Kendall and Pearson coefficients between the MIR of any labelled input $x$ and the error rate of transductive learning (\% of unlabelled inputs incorrectly inferred by the SSL algorithm) after poisoning the label of $x$ only. These coefficients take their value between -1 and +1 (negative and positive correlations). Coefficient values greater than 0.5 demonstrate a strong correlation; between 0.3 and 0.49 correspond to medium correlations; less than 0.29 are interpreted as weak correlations.

To perform transductive learning, we consider both label propagation and label spreading. This allows us to observe how well our methods transfers to another SSL algorithm computing input influences differently.

We measure the correlation for all datasets, with different proportions of labelled/unlabelled inputs, i.e., with 5\%, 15\% and 25\% of labelled inputs (see Table \ref{tab:rq1} for exact numbers for each dataset). We start from 5\% because this is the smallest percentage where the SSL algorithms yield acceptable accuracy values (above 80\%) for all three datasets. At the opposite end, 25\% is aligned with current research on SSL (see, e.g. \cite{PapersSSL}). 
Considering different proportions of labelled inputs allows us to observe whether the MIR metric keeps the same effectiveness when an increasingly smaller set of unlabelled inputs are inferred from an increasingly larger set of labelled inputs. For each proportion, we randomly select which inputs are labelled. To account for random variations resulting from this split, we repeat each experiment 3 to 10 times (depending on the computation cost involved) and report the average of the correlation coefficients.

\subsubsection{Results} 

\begin{table}
    \centering                      
    \begin{tabular}{r||rr|rr|rr|}
         \textbf{MNIST} & \multicolumn{2}{c|}{$l = 650$ (5\%)} & \multicolumn{2}{c|}{$l = 1950$ (15\%)} & \multicolumn{2}{c|}{$l = 3250$ (25\%)} \\
         \hline
         L. propagation & 0.96 & 0.99 & 0.95 & 0.97  & 0.95 & 0.98 \\
         L. spreading & 0.62 & 0.83 & 0.63 & 0.80 & 0.59 & 0.71   \\
        \multicolumn{7}{c}{}\\
         \textbf{CIFAR-10} & \multicolumn{2}{c|}{$l = 500$ (5\%)} & \multicolumn{2}{c|}{$l = 1500$ (15\%)} & \multicolumn{2}{c|}{$l = 2500$ (25\%)} \\
         \hline
         L. propagation & 0.42 & 0.52 & 0.43 & 0.49  & 0.41 & 0.47 \\
         L. spreading & 0.35 & 0.35 & 0.31 & 0.32 & 0.30 & 0.33   \\
         \multicolumn{7}{c}{}\\
         \textbf{RCV1} & \multicolumn{2}{c|}{$l = 1012$ (5\%)} & \multicolumn{2}{c|}{$l = 1036$ (15\%)} & \multicolumn{2}{c|}{$l = 5060$ (25\%)} \\
         \hline
         L. propagation & 0.47 & 0.73 & 0.51 & 0.81  & 0.56 & 0.82 \\
         L. spreading & 0.42 & 0.53 & 0.43 & 0.53 & 0.43 & 0.57   \\

    \end{tabular}
    \caption{Correlation, across all labelled inputs, between the MIR metric of any input and the error rate of transductive learning after altering the input label. Left numbers are Kendall's $\tau$ coefficients while right numbers are Pearson's. 
    }
    \label{tab:rq1}
\end{table}

\textbf{General observations.} As observed in Table \ref{tab:rq1}, the Kendall and Pearson coefficients show medium to strong correlations in all experiments (Kendall coefficients are between 0.30 and 0.96, while Pearson coefficients are between 0.32 and 0.99). All reported correlation coefficients are associated with a p-value less than $10
^{-7}$, meaning that we can reject the null hypothesis of an overall absence of correlation with a type-1 error $<1\%$. Overall, this means that \emph{our MIR metric captures well the estimated impact of poisoning any labelled input}. Thus, MIR has the potential to prioritize the inputs that an attacker should target to maximize error rate (which we assess in RQ2 and RQ3). Moreover, the high Pearson coefficients demonstrate that MIR also enables a relative comparison of the input impact (i.e., a doubled MIR indicates a doubled increase in error rate).

\textbf{Dataset.} The correlation values vary much across the datasets. This can be explained by the fact that SSL heavily relies on the relative position of the class clusters during label inference. Thus, label poisoning is more efficient as the clusters are closer (the influence of the negative class is stronger on the positive inputs). For instance, MNIST inputs (white digits on dark background) are closer in the feature space than CIFAR-10 inputs (coloured images with arbitrary background), which explains that the correlations in MNIST are stronger than in CIFAR-10.

\textbf{Transductive learning algorithm.} Stronger correlations are observed when transductive learning is performed by label propagation. This was expected since our approach follows the same energy propagation principle as label propagation does. This reveals that the direct influence of any labelled input on any unlabelled input is a good approximation of its total influence (including indirect propagation through intermediary inputs). When label spreading is used instead, the correlations are slightly lower but remains within the same range of correlation strength. This indicates that \emph{the MIR metric transfers well over different graph-based SSL algorithms}. For engineers, this means that relying on one algorithm over the other increases poisoning resilience only slightly.

\textbf{Proportion of labelled inputs.} The relative proportion of labelled inputs (compared to unlabelled inputs) does not significantly affect the correlations.  Interestingly, this means that \emph{MIR remains a strong metric even when there are more labelled inputs competing to influence unlabelled inputs}. In other words, (manually) labelling more inputs does not hinder the capability of MIR to select the most impactful inputs.

\subsection{RQ2: Impact on Inductive Error Rate}


\subsubsection{Detailed setup}

We measure the effectiveness of our poisoning method by computing the error rate of inductive learning (using a Random Forest -- RF -- and a Multi-Layer Perceptron -- MLP). We label 25\% of inputs.\footnote{Because the proportion of labelled inputs does not affect our conclusions and because of lack of space, we do not report here the results for 5\% and 15\%. For the sake of reproducibility, the results for 5\% and 15\% of labelled inputs can be computed with our replication package.} Then, we poison the $k$ labelled inputs with the highest MIR, for $k$ ranging from $5\%$ to $20\%$ of the labelled inputs (by steps of $5\%$), and we measure the error rate. Compared to RQ1, these experiments allow us to measure (a) the effectiveness of our method when selecting a set of inputs (rather than a single one) and (b) how much the attack transfers during induction.

To perform the inductive learning, we first apply an SSL algorithm (i.e., each amongst label propagation and label spreading) to infer the labels of the unlabelled input set. Then, we train a supervised model (i.e. , an RF or an MLP) using both labelled inputs (with their known labels) and the unlabelled inputs (with their inferred labels). This yields, per dataset, a total of four combinations (2 SSL algorithms $\times$ 2 supervised models).


We show the inductive error rate for the four combinations, on all datasets, with a fixed proportion of labelled/unlabelled inputs (i.e., 25\% of labelled inputs in the training set), and with different budget $k$ of those labelled inputs that our method can poison (5\%, 10\%, 15\% and 20\% of the labelled inputs). For instance, with 25\% of labelled inputs in CIFAR-10, a 5\% poison budget means that our method has poisoned the 250 labelled inputs with the highest MIR out of the 5000 labelled inputs.

To account for random variations when splitting the training set into labelled and unlabelled inputs, we repeat each experiment five times and report the average of the obtained inductive error rates.

\begin{figure*}
    \centering
        \includegraphics[width=0.32\linewidth]{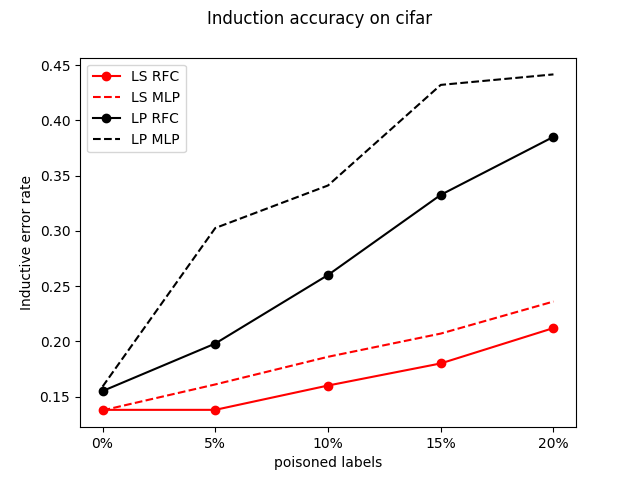} 
        \includegraphics[width=0.32\linewidth]{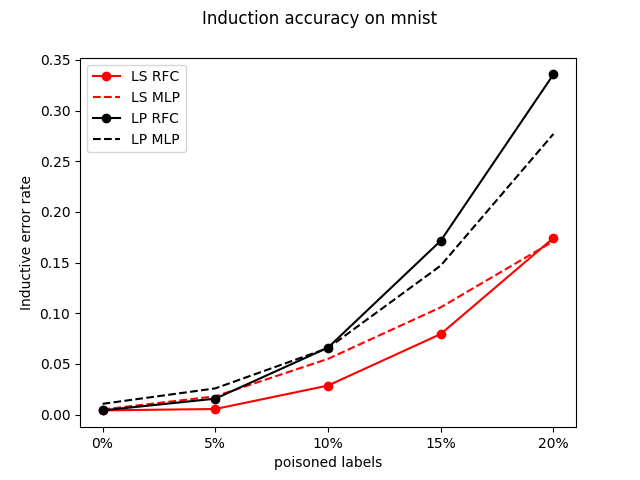}
        \includegraphics[width=0.32\linewidth]{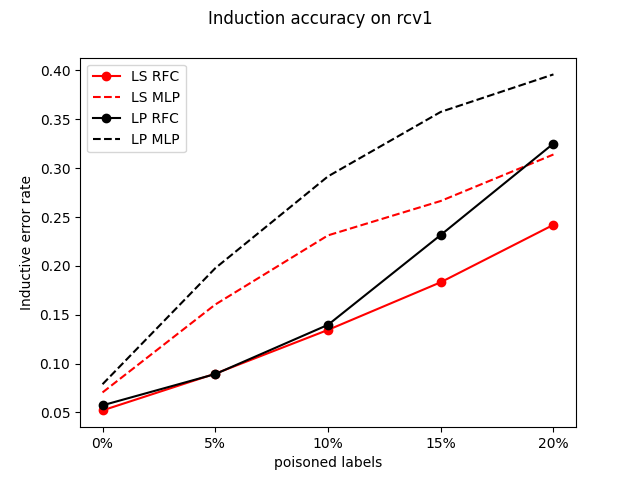}
    \caption{Error rate (Y axis) achieved by our data poisoning attack wrt. an increasing poison budget (X axis, expressed in \% of the labelled inputs), measured on random forest classifier (RFC plain curves) and multilayer perceptron (MLP dotted curves) trained on both known labels and labels inferred with label propagation (LP black curves) or label spreading (LS red curves).}
    \label{fig:rq2}
\end{figure*}

\subsubsection{Results} \textbf{General observations.} As shown in Figure~\ref{fig:rq2}, our poisoning attack significantly increases the error rate of the supervised models, regardless of the SSL algorithm used for transductive learning. For example, when poisoning 20\% of the labelled inputs, the total increase in error rate ranges from 16\% (MNIST, label spreading, RF) to 45\% (CIFAR-10, label propagation, MLP). These results indicate that \emph{(1) the MIR metric remains effective when multiple inputs are poisoned and (2) the induced errors transfer to inductive learning}. This second point demonstrates that introducing a different learning algorithm in the induction process does not impede the altered labels to ``poison the well''.

\textbf{SSL algorithm.} The choice of the SSL algorithm to infer the labels of the unlabelled inputs has a major effect on the transferability of MIR-based data poisoning. Indeed, using label spreading alleviates not only the total error rate but also the marginal effect of increasing the poison budget. For instance, we observe in left figure of Figure~\ref{fig:rq2} (which concerns CIFAR-10) that the error rate increase for both supervised models is twice steeper when using label propagation than when using label spreading. This trend is also observed on MNIST and RCV1, where the error rate curve is steeper in the case of label propagation. These results shed new light on our findings from RQ1: \emph{while the choice of the SSL algorithm does not change the impact of the first poisoned inputs on transductive learning, it allows increasing the resilience of inductive learning when more inputs are poisoned}.

\subsection{RQ3: Comparison with the State of the Art}

\subsubsection{Detailed setup}
We compare the effectiveness our method to the state of the art (i.e., the \texttt{greedy} method and the \texttt{probabilistic} method of Liu et al. \cite{liu2019}), in terms of effectiveness (error rate of transductive learning and inductive learning) and efficiency (computing time required to select the inputs to poison). 

To compare efficiency, we measured the time required by each method to select a subset of 20 labelled inputs to poison, on all datasets. For our method, this involves computing the MIR value of all labelled inputs and rank them accordingly. 
For the greedy method, this means performing 20 iterations of their greedy search algorithm. For the probabilistic method it means estimating the poisoning probability for each labelled samples based on their objective function and returning the most probable 20 samples.  In case the poisoning budget exceeds 20 inputs, the greedy method takes proportionally more time, whereas the computing time of our method and of the probabilistic one are not affected by the number of poisoned inputs. To account for random variations, we repeat the experiments 50 times and report the median and the standard deviation.

To compare effectiveness, we apply the same experimental protocol as RQ2, with the addition that we also compute the error rate of label propagation during transductive learning.

\subsubsection{Results}

\begin{table}
    \centering

    \begin{tabular}{r||c|c|c|c|c|c}
        \toprule
        \textbf{} & \multicolumn{2}{c}{MNIST} & \multicolumn{2}{|c|}{RCV1} &
        \multicolumn{2}{c}{CIFAR-10} \\
        \hline
        &$\mu$&$\sigma$&$\mu$&$\sigma$&$\mu$&$\sigma$ \\
        \midrule
        \textbf{Greedy} & 86.47           & 4.58               & 711.25  & 77.69 & 479.02 & 11.05              \\
        \textbf{Probabilistic} & 29.65           & 0.80               & 109.51  & 0.82 & 17.52 & 0.71              \\
        \textbf{MIR}   & 7.67            & 0.19              & 55.45   & 0.38 & 6.44 & 0.75 \\
    \end{tabular}
    \caption{Efficiency of our method (MIR) compared to the state of the art (greedy and probabilistic). $\mu$ and $\sigma$ are the median and the standard deviation of the runtime (in s.).}
    \label{tab:rq3-efficiency}
\end{table}

\textbf{Efficiency.} The median and standard deviation of each method's runtime on each dataset is given in Table \ref{tab:rq3-efficiency}. We observe that our method significantly outperforms the other two in every case, being 2 to 3 times faster than the probabilistic method and an order of magnitude faster than the greedy method, while having a smaller standard deviation. 

\textbf{Effectiveness.} The error rates achieved by the three methods are shown in Tables~\ref{tab:rq3-effectiveness}. When applied to transductive learning, our MIR-based method achieves a higher error rate than the other two for every dataset and poison budget. 
The difference increases as more inputs are poisoned. This indicates that our method exploits additional poison budget better than the other methods. 

When applied to inductive learning, the general trends remain: our method generally outperforms the others and it does so better as the poison budget increases. For MNIST and CIFAR-10, the difference is substantial: our method achieves an error rate 1.22 to 30.8 times higher. 

As for RCV1, the three methods perform comparatively well, though the probabilistic method works best when a RF is used for induction. A Wilcoxon signed rank test reveals that the difference between our method and the probabilistic one (over all RCV1 results) is not statistically significant. Nonetheless, by extrapolating from the efficiency results (Table \ref{tab:rq3-efficiency}), we estimate that MIR would run 17 times faster on RCV1 than the deterministic method already with a poison budget of 5\% (even more as the budget increases).

Taken together, all our results show that our method results in a higher error rate than the state of the art, with statistical significance (the Wilcoxon test over all results rejects the null hypothesis that the two methods perform equally, with a p-value less than $10^{-11}$). 
Thus, \emph{our influence-driven attack  outperforms the state of the art in both effectiveness and efficiency and forms a new baseline for SSL poisoning}.

\begin{table*}
    \centering
    \begin{tabular}{c||c|ccc|ccc|ccc|ccc}
        \hline
         & clean & \multicolumn{3}{c|}{5\% labels poisoned} & \multicolumn{3}{c|}{10\% labels poisoned} & \multicolumn{3}{c|}{15\% labels poisoned}
         & \multicolumn{3}{c}{20\% labels poisoned}\\
         \hline
         \hline
         \textbf{Trans.}  & & MIR & Greed. & Prob.  & MIR & Greed. & Prob. & MIR & Greed. & Prob.  & MIR & Greed. & Prob. \\
         \hline
         MNIST & & & & & & & & & & & & &\\
         5\% labels & 1.25\% & \textbf{16.22\%} &3.04\%& 5.41\% & \textbf{27.05\%} &3.04\%&9.69\%  & \textbf{35.86\%} &3.04\%&15.13\%  & \textbf{43.98\%}&3.04\%&19.46\% \\
         15\% labels & 0.69\% &\textbf{16.09\%} & 1.27\%&5.67\% & \textbf{27.48\%} &1.27\%&9.84\% &\textbf{37.07\%} &1.27\%&14.71\% &\textbf{45.71\%} &1.27\% & 20.17\%\\ 
         25\% labels &0.58\% &\textbf{17.47\%} &1.16\%& 5.68\%  &\textbf{29.94\%} &1.16\%&9.86\% &\textbf{40.26\%} &1.16\%&14.82\% &\textbf{48.88\%} &1.16\%&19.26\% \\ 
         \hline
         CIFAR-10 & & & & & & & & & & & & &\\
         5\% labels &19.12\% &\textbf{33.62\%} & 26.15\%&21.90\% &\textbf{53.08\%}& 28.90\% &25.20\% &\textbf{61.67\%}  &28.90\%&25.00\%  &\textbf{66.29\%} &28.90\%& 31.90\% \\
         15\% labels &17.52\%&\textbf{32.03\%} &26.62\%&18.70\% &\textbf{37.98\%} &26.62\%&19.10\% &\textbf{45.00\%} &26.62\%&26.01\% &\textbf{51.66\%} &26.62\%&28.90\%\\ 
         25\% labels &17.34\%&\textbf{29.21\%} &28.11\%&19.40\% &\textbf{35.67\%} &28.99\%&20.80\% &\textbf{40.21\%} &28.99\% &19.20\% &\textbf{44.20\%} &28.99\%&21.20\% \\ 
         \hline
         
         RCV1 & & & & & & & & & & & & &\\
         5\% labels &11.66\% &\textbf{22.22\%} &19.53\% &19.66\% &\textbf{29.39\%}& 28.62\% &28.72\% &\textbf{35.17\%} &34.62\% &35.02\% &\textbf{42.37\%} &38.77\%&39.85\%\\
         15\% labels &8.19\% &\textbf{22.00\%} &18.13\%&18.69\% &\textbf{31.26\%} &27.09\%&28.01\% &\textbf{38.89\%} &34.74\% &35.28\% &\textbf{46.26\%} &39.10\%&39.96\%\\ 
         25\% labels &6.91\% &\textbf{23.12\%} &18.41\% &19.31\% &\textbf{33.18\%} &27.98\% &28.22\% &\textbf{41.32\%} &34.62\%&35.15\% &\textbf{49.74\%} &39.57\% &40.12\%\\ 
         \hline
         \hline
         
         \textbf{Ind. R.F.}  & & MIR & Greed. & Prob.  & MIR & Greed. & Prob. & MIR & Greed. & Prob.  & MIR & Greed. & Prob. \\

         \hline
         MNIST & & & & & & & & & & & & &\\
         5\% labels &1.07\% &\textbf{8.93\%} &1.25\%&0.92\% &\textbf{19.94\%} &1.25\%&1.34\%  &\textbf{29.22\%} &1.25\%&2.64\% &\textbf{40.92\%} &1.25\% &4.35\%\\
         15\% labels &0.56\% &\textbf{2.27\%} &1.56\%&0.55\% &\textbf{10.31\%} &1.56\%&0.55\%  &\textbf{19.61\%}&1.56\%&0.83\% &\textbf{34.17\%} &1.56\%&1.66\% \\ 
         25\% labels &0.42\% &\textbf{1.62\%} &0.51\%&0.55\% &\textbf{6.85\%} &1.43\%&0.42\% &\textbf{18.50\%} &1.43\%&0.60\% &\textbf{33.71\%} &1.43\%&1.57\% \\ 
         \hline
         CIFAR-10 & & & & & & & & & & & & &\\
         5\% labels &18.96\% &22.36\% &\textbf{25.35}\%&17.60\% &\textbf{53.32\%} &28.81\%&19.40\% &\textbf{68.81\%} &28.81\%&20.15\% &\textbf{71.76\%} &28.81\%&21.55\% \\
         15\% labels &17.25\% &22.10\% &\textbf{24.06}\%&18.55\% &\textbf{29.1\%} &24.04\%&19.25\% &\textbf{43.32\%} & 24.04\%&20.65\% &\textbf{53.45\%} &24.04\%&23.75\% \\ 
         25\% labels &15.50\% &\textbf{20.55\%} &18.95\%&15.10\% &\textbf{25.70\%} &20.19\%&15.80\% &\textbf{29.60\%} &20.19\%&17.85\% &\textbf{39.90\%} &20.19\%&20.05\% \\ 
         
         \hline
        RCV1 & & & & & & & & & & & & &\\
5\% labels &9.00\% &11.44\% &10.49\%&\textbf{13.79\%}  &15.96\% &16.45\%&\textbf{36.56}\% &23.09\% &26.36\%&\textbf{48.64\%} &35.92\% &29.36\%&\textbf{52.00\%} \\
         15\% labels &5.85\% &9.77\% &10.85\%&\textbf{13.28\%} &15.33\% &14.22\%&\textbf{38.88\%} &26.84\% &23.38\%&\textbf{51.16\%} &44.08\% &23.38\%&\textbf{53.40\%} \\ 
         25\% labels &5.73\% & 8.92\% &11.64\%&\textbf{11.68\%} & 13.96\% &20.88\%&\textbf{33.60\%}& 23.17\% &30.08\%&\textbf{50.40\%}&32.60\%& 30.08\%&\textbf{53.40\%} \\

        \hline
        \hline
        
         \textbf{Ind. MLP}  & & MIR & Greed. & Prob.  & MIR & Greed. & Prob. & MIR & Greed. & Prob.  & MIR & Greed. & Prob. \\
         \hline
         MNIST & & & & & & & & & & & & &\\
5\% labels & 2.17\%&\textbf{10.17}\%&3.42\%&2.31\%&\textbf{20.76}\%&3.42\%&2.40\%&\textbf{33.15}\%&3.42\%&5.41\%&\textbf{45.31}\%&3.42\%&7.12\% \\
15\% labels &0.88\%&\textbf{4.53}\%&1.06\%&1.20\%&\textbf{16.18}\%&1.06\%&1.02\%&\textbf{25.98}\%&1.06\%&1.39\%&\textbf{42.07}\%&1.06\%&2.22\% \\
25\% labels &0.92\%&\textbf{4.58}\%&1.06\%&0.69\%&\textbf{7.17}\%&1.06\%&0.92\%&\textbf{14.33}\%&1.06\%&1.06\%&\textbf{22.47}\%&1.06\%&1.99\% \\
        
        \hline
         CIFAR-10 & & & & & & & & & & & & &\\
         5\% labels &19.00\%&\textbf{21.85}\%&21.80\%&18.80\%&\textbf{64.30}\%&20.15\%&29.70\%&\textbf{66.45}\%&20.15\%&26.90\%&\textbf{69.80}\%&20.15\%&23.90\%\\
15\% labels &16.85\%&19.05\%&\textbf{31.20}\%&18.10\%&24.25\%&\textbf{32.15}\%&18.60\%&\textbf{44.15}\%&32.15\%&18.05\%&\textbf{51.30}\%&32.15\%&20.25\%\\
25\% labels &16.50\%&\textbf{17.60}\%&15.95\%&16.90\%&\textbf{19.65}\%&16.85\%&17.45\%&\textbf{27.70}\%&16.85\%&17.30\%&\textbf{40.50}\%&16.85\%&19.65\%\\
         
         \hline
         RCV1 & & & & & & & & & & & & &\\
5\% labels&11,68\%&17,88\%&21,16\%&\textbf{21,52}\%&26,56\%&\textbf{32,00}\%&31,72\%&32,56\%&\textbf{41,08}\%&40,68\%&39,68\%&46,24\%&\textbf{46,92}\%\\
15\% labels&7,40\%&\textbf{19,16}\%&18,92\%&19,00\%&29,08\%&\textbf{32,92}\%&30,04\%&35,88\%&40,76\%&\textbf{41,28}\%&40,48\%&45,04\%&\textbf{46,56}\%\\
25\% labels&6,72\%&\textbf{21,12}\%&18,96\%&20,28\%&\textbf{29,48}\%&27,64\%&27,44\%&35,32\%&33,68\%&\textbf{36,56}\%&40,80\%&40,08\%&\textbf{42,64}\%\\
    \hline
    \end{tabular}
    \caption{RQ3: Error rate (in \%) of \emph{transductive learning} (with label propagation), \emph{inductive learning} (with random forest) and \emph{inductive learning} (with multi-layer perceptron) achieved by our MIR method (MIR), the state of the art deterministic (Greed) and stochastic (Prob) methods. Higher is better. Bold indicates the best method for each experiment.}
    \label{tab:rq3-effectiveness}
\end{table*}

\subsection{RQ4: Countermeasures}

\subsubsection{Detailed setup}
We consider again the cases of transductive learning and inductive learning, where 5\%, 15\%, and 25\% of the inputs are labelled. We apply our influence-driven attack with a poison budget totalling 10\% of the labelled inputs. Then, we investigate how much the two countermeasures (relabelling inputs with the highest MIR vs. labelling additional (previously unlabelled) inputs). We allocate the same effort to the two countermeasures, i.e., one third of the number of poisoned labels. We repeat the experiments five times and report the average error rates.

\subsubsection{Results}
Table \ref{tab:rq4} shows the results. We observe that relabelling the most influential labels systematically achieves a higher reduction in error rate. Interestingly, on average, relabelling one third of the poisoned inputs halves the error rate induced by the poisoning. A Wilcoxon test rejects the null hypothesis that the two methods perform equally (p-value $< 10^{-5}$). Labelling more input actually offers small reductions in error rate ($< 15\%$ on average). Overall, this indicates that \emph{to alleviate the poisoning effects, engineers should focus their effort on relabelling influential inputs}.



\begin{table}
    \centering
    \begin{tabular}{|c||ccc|ccc|}
    \hline
         & \multicolumn{3}{c|}{Transduction} & \multicolumn{3}{c|}{Induction}\\
         \textbf{MNIST} & MIR & LAB & None & MIR & LAB & None\\
         \hline
         5\% labels & \textbf{15.80} & 24.06 & 27.05 & \textbf{04.38} & 12.32 & 19.94 \\
         15\% labels & \textbf{16.38} & 25.64 & 27.48 & \textbf{01.56} & 06.02 & 10.31 \\
         25\% labels & \textbf{17.12} & 27.21 & 29.94 &\textbf{01.00} & 03.80 & 6.85 \\
         \hline
    \end{tabular}\\
    \vspace{0.2cm}   
    \begin{tabular}{|c||ccc|ccc|}
    \hline
         & \multicolumn{3}{c|}{Transduction} & \multicolumn{3}{c|}{Induction}\\
         \textbf{CIFAR-10} & MIR & LAB & None & MIR & LAB & None\\
         \hline
         5\% labels & \textbf{28.76} & 46.06 & 53.08 & \textbf{19.46} & 42.39 & 54.95\\
         15\% labels & \textbf{23.17} & 36.35 & 37.98 & \textbf{19.47} & 43.44 & 54.11\\
         25\% labels & \textbf{23.75} & 34.78 & 35.67 & \textbf{17.18} & 41.36 & 44.06\\
         \hline
    \end{tabular}\\
    \vspace{0.2cm}
    \begin{tabular}{|c||ccc|ccc|}
    \hline
         & \multicolumn{3}{c|}{Transduction} & \multicolumn{3}{c|}{Induction}\\
         \textbf{RCV1} & MIR & LAB & None & MIR & LAB & None\\
         \hline
         5\% labels & \textbf{21.64} & 28.00 & 29.39 & \textbf{10.56} & 13.92 & 15.96\\
         15\% labels & \textbf{19.58} & 29.62 & 31.26 & \textbf{09.04} & 14.04 & 15.33\\
         25\% labels & \textbf{20.31} & 31.07 & 33.18 & \textbf{08.20} & 12.24 & 13.96\\
         \hline
    \end{tabular}\\
    \caption{RQ4: Error rate (in \%) achieved by double checking one third of the poisoned labels (MIR) compared with labelling the same number of additional inputs (LAB) and no countermeasure (None). Lower is better.}
    \label{tab:rq4}
\end{table}

\subsection{Threats to Validity}

Threats to internal validity concern the implementation of the software artefacts used in our study. 
Some are addressed by the fact that we reuse established implementations of the learning algorithms with typical parameters. The resulting (non-poisoned) models yield a small error rate on state-of-the-art datasets used as is (including their splitting into training and test sets), even when small proportions of labelled inputs are used. This indicates that our setup was appropriate. 

The implementation of our approach was tested manually and through various experiments, which provides some confidence regarding its correctness. Moreover, we reused the available implementation of the greedy and the probabilistic method and as provided by its inventors.

The threats to external validity originate from the number of learning algorithms and datasets we used in our experiments. MNIST and CIFAR-10 are established in the ML literature, whereas RCV1 was already used in SSL-related studies \cite{liu2019}. The reduction to binary classification problems has followed the protocol of previous research \cite{liu2019,LIBSVMdata}.

The randomness induced by splitting the training set between labelled inputs and unlabelled inputs is another factor affecting our results. To mitigate its effects, we repeated our experiments multiple times (between 3 and 10, depending on required the computation time) and manually checked the absence of significant variations. Still, it remains possible that additional runs produce outliers (especially for RCV1, the largest dataset of the three).

Nevertheless, in general, it is likely that the effectiveness of our approach varies upon different external factors such as the used datasets (as our experiments already witness). Still, the core principle of measuring input influence is universally applicable and we believe that our key conclusions shall remain valid. Only additional experimentations can alleviate this risk, though. Fortunately, our open-source implementation and the black-box nature of our influence metric facilitate the replication and complementation of our study.

Finally, threats to construct validity come from the factors we measure to draw our conclusions. We studied the correlation between MIR (our metric) and the error rate, which is a natural metric to use. The error rate was also used by previous research \cite{liu2019} to measure the effectiveness of data poisoning.

\section{Conclusion}

We proposed a new label poisoning attack that targets the most influential inputs during semi-supervised learning. Our extensive experiments show that our approach outperforms the state of the art (in effectiveness and efficiency) under various settings and forms a new baseline for future research. 

The influence metric it leans on has the advantage of being simple and fast to compute. In addition to allowing the design of efficient and effective attacks, it also provides practical benefits for engineers. Indeed, using the same simple metric, engineers can identify the most critical inputs to investigate in case there is doubt about data integrity (e.g., when data come from untrustworthy sources). This way, they can reduce the risk and effects of poisoning by relabelling influential inputs and, overall, alleviate the data quality threats.

\section*{Acknowledgement}

This work is supported by the Luxembourg National Research Funds (FNR) through CORE project C18/IS/12669767/STELLAR/LeTraon.

\balance
\bibliographystyle{ACM-Reference-Format}
\bibliography{references}

\end{document}